\documentclass[letterpaper, 10 pt, conference]{ieeeconf}

\usepackage{graphicx}
\usepackage{amsmath,amssymb}
\usepackage{siunitx}
\usepackage{upgreek}
\usepackage[noend]{algpseudocode}
\usepackage{algorithm}
\usepackage{multirow}
\usepackage{url}
\usepackage{balance}
\usepackage{xcolor}
\usepackage{hyperref}
\usepackage{caption}
\usepackage{subcaption}
\usepackage{multicol}

\begin{document}

\title{\LARGE \bf GESD: Beyond Outcome-Oriented Fairness}

\author{Gideon Popoola and John Sheppard\\
\textit{Montana State University} \\
 Bozeman, MT, USA \\
 gideon.popoola@student.montana.edu, john.sheppard@montana.edu}
\thanks{$^{1}$Gideon Popoola is a PhD student with the Gianforte School of Computing, Montana State University, Bozeman, MT 59717, USA {\tt\small gideon.popoola@student.montana.edu}}%
\thanks{$^{2}$John Sheppard is a Professor with the Gianforte School of Computing, Montana State University, Bozeman, MT 59717, USA
        {\tt\small john.sheppard@montana.edu}}%

\maketitle

\begin{abstract}
Machine learning (ML) algorithms are increasingly deployed in high-stakes decision-making domains such as loan approvals, hiring, and recidivism predictions. While existing fairness metrics (e.g., statistical parity, equal opportunity) effectively quantify outcome-oriented disparities, they offer limited insight into the procedure or explanation behind biased decisions. To address this gap, we propose Group-level Explanation Stability Disparity (GESD), a \textit{procedural-oriented} fairness metric that measures disparities in the stability, robustness, and sensitivity of model explanations across different subgroups in a protected category. 
We further integrate GESD into a multi-objective optimization framework that jointly optimizes for utility, outcome-based fairness, and explanation-based fairness called FEU (Fairness--Explainability--Utility). Empirical results on multiple benchmark datasets show that GESD effectively captures group-wise discrepancies in explanation quality, and that FEU improves both utility and fairness over state-of-the-art methods. By bridging outcome-based and explanation-based fairness, GESD offers a comprehensive tool for diagnosing and mitigating bias in predictive modeling. Our code and datasets are available on GitHub {\hyperlink{https://github.com/horlahsunbo/GESD}{https://github.com/horlahsunbo/GESD}}
\end{abstract}

%

\section{Introduction}
\label{sec:intro}
Machine learning (ML) algorithms are increasingly deployed in high-impact domains such as hiring, college admissions, loan approvals, and criminal justice \cite{barocas2016big}. Despite the potential for efficiency and scalability, previous works have raised serious concerns about algorithmic bias, where models are systematically unfair to certain members of protected categories such as age, race, and gender \cite{ popoola2024investigating}. Early research determined that ML systems can inherit social prejudices from the datasets they are trained on, inadvertently amplifying historical biases \cite{calmon2017optimized}. These findings led to developing a series of outcome-oriented fairness metrics, which focus on measuring and mitigating disparities in model prediction or outcomes, such as statistical parity \cite{chouldechova2017fair}, equalized odds \cite{hardt2016equality}, and predictive parity \cite{corbett2017algorithmic}.

Although these outcome-oriented metrics can quantify bias and optimize the model for fairness, most of these methods still rely on model predictions. This reliance on model prediction means these metrics ignore the potential bias in decision-making (otherwise known as the procedure or explanation). This negligence means the metric cannot further mitigate the procedural-oriented bias, which is crucial for developing a fair and explainable ML model \cite{alvarez2018robustness}.


Given the above, there is a growing research community interested in bridging the gap between explainability, which represents the degree to which humans can understand or trust the reasoning of a model, and fairness, which represents the extent to which the model does not unfairly disadvantage protected categories. Lipton (\cite{lipton2018mythos}) highlights that merely observing fair outputs might be insufficient, which means an algorithm can produce seemingly equitable outcomes but still rely on questionable or discriminatory internal procedures. Recent studies have pointed to the necessity of examining model procedures rather than relying solely on outcome statistics \cite{grabowicz2022marrying}. Some efforts propose examining how sensitive the reasoning of a model is when masking the most important feature, known as explanation fidelity \cite{alvarez2018towards}. However, these remain limited in scope because they still rely somewhat on model prediction. A more comprehensive, procedural-oriented fairness metric would explicitly evaluate whether members of different protected attributes receive explanations with similar degrees of stability, robustness, and sensitivity. Such a metric could, for example, detect if the explanatory patterns of a model drastically fluctuate across subgroups.

Beyond bridging fairness and explainability, recent studies have also begun treating fairness, explainability, and utility as a multi-objective problem in machine learning models. One common approach to addressing this problem is through the use of a composite loss function, which combines metrics 
into a single loss function with tunable weights \cite{zhao2023fairness}.
Furthermore, Pareto-based strategies explore the spectrum of balancing the objectives using methods like multi-objective evolutionary algorithms (MOEA) to generate Pareto sets that balance utility, explainability, and fairness \cite{wang2024generating}.


This paper introduces a procedural-oriented fairness framework to balance fairness and explainability. Specifically, we define the Group-level Explanation Stability Disparity (GESD) metric to quantify the robustness, stability, and sensitivity of model explanations across protected subgroups, thereby exposing procedural biases that outcome-oriented fairness metrics often overlook. Unlike prior measures that compare average feature importance or mean explanations across groups \cite{zhao2023fairness}, GESD evaluates the reliability of explanations themselves by assessing their stability under perturbations, ensuring that different groups receive explanations of comparable consistency. The metric is both model- and explainer-agnostic, making it compatible with a wide range of learning algorithms and post-hoc interpretability methods. Importantly, GESD complements traditional prediction-level fairness metrics by identifying disparities in explanation quality even when a model appears fair in terms of its outcomes. 

Under our framework, we integrate GESD into a multi-objective optimization process called Fairness-Explanation-Utility (FEU), which jointly optimizes outcome-based fairness, explanation-based fairness, and model utility. Our approach thus provides a more holistic view of equitable modeling, ensuring that both the final predictions and the underlying process are fair across protected groups. The contributions of this work are summarized as follows:
\begin{enumerate}
    \item We formulate the problem of detecting group-level disparities in explanation quality and propose a new procedural-oriented metric (GESD) to quantify disparities in explanation robustness, stability, and sensitivity.
    \item We proposed an evolutionary algorithm-inspired multi-objective optimization framework named FEU that optimizes utility, fairness, and explainability metrics to generate a balanced model in these objectives.
    \item We demonstrate that GESD effectively detects hidden group-level explanation bias on multiple benchmark datasets. 
    \item We show that FEU can achieve superior trade-offs in utility and fairness compared to existing fair models.
\end{enumerate}

\section{Group-level Explanation Stability Disparity}
\label{sec:gesd}
GESD 
builds on prior definitions of explanation sensitivity and robustness to ensure that no group receives significantly less stable explanations than another \cite{montavon2018methods, alvarez2018robustness}. We 
show the complete algorithm in Algorithm ~\ref{algorithm1}.

\begin{algorithm}[t!]
\caption{Group-level Explanation Disparity (GESD)}\label{algorithm1}
\begin{algorithmic}[1]
\Require Trained model $f$, evaluation dataset $X \in \mathbb{R}^{n \times d}$, sensitive attributes $s \in \mathbb{R}^n$, training data $X_{\text{train}}$, number of perturbations $N$, Gaussian scale $\sigma$, mask probability $p$, baseline value $b$, sample size $m$
\State \textbf{Initialize} stability list $\mathcal{S} \gets \emptyset$
\State Randomly sample $X_{\text{sample}} \subset X$ of size $m$
\For{each $x \in X_{\text{sample}}$ with sensitive attribute $s_x$}
    \State Compute $E_{\text{SHAP}}(x)$ using $f$ and $X_{\text{train}}$
    \State Compute $E_{\text{LIME}}(x)$ using $f$ and $X_{\text{train}}$
    \State Compute aggregated explanation (Eq. \ref{eq:centagg})
    \State \textbf{Initialize} perturbation stability list $L \gets \emptyset$
    \For{$i=1$ to $N$}
        \State Generate noise: $\epsilon \sim \mathcal{N}(0, \sigma^2 I)$
        \State Compute perturbed instance: $x' \gets x + \epsilon$
        \State Generate binary mask: \[M \in \{0,1\}^d,\ p_j=P(M_j = 0),\  
        j=1,\ldots,d\]
        \State Apply masking: 
        \[
        \tilde{\mathbf{x}} \gets M \odot (\mathbf{x} + \boldsymbol{\epsilon}) + (\mathbf{1}-M)\odot \mathbf{x}^{(0)}
        \]
        \State Compute perturbed instance explanation (Eq. \ref{eq:centagg})
        \State Compute distance (Eq. \ref{eq:instab})
        
        \State Compute stability score for the perturbation:
        \[
        S(x) \gets \frac{1}{1+\Delta{(x)}}
        \]
        \State Append $S_i(x)$ to $L$
    \EndFor
    \State Compute average stability for $x$:
    \[
    S_i(x) \gets \frac{1}{N}\sum_{i=1}^{N} S(x)
    \]
    \State Record the pair $(s_x, S(x))$ in $\mathcal{S}$
\EndFor
\For {each group $G_i$} 
\[
S_i \gets \frac{1}{|X_{G_i}|} \sum_{\mathbf{x}\in X_{G_i}} S_i (\mathbf{x})~,
\]
\EndFor
\If{number of groups $K = 2$}
    \[
    \text{GESD} \gets \left|{S}_{0} - {S}_{1}\right|
    \]
\Else
\begin{eqnarray*}
    \bar{S} &\gets& \frac{1}{K}\sum_{g=1}^{K} {S}_i\\
    \text{GESD}_{K\text{-groups}} &\gets& \frac{1}{K}\sum_{g=1}^{K} \left({S}_i - \bar{S}\right)^2
\end{eqnarray*}
\EndIf
\State \Return \text{GESD}
\end{algorithmic}
\end{algorithm}

\subsubsection{Explanation Aggregation}
Let $f(\mathbf{x})$ be the prediction of a model for input $\mathbf{x}\in\mathbb{R}^d$. Let $E_{\text{SHAP}}(f,\mathbf{x}) = (\phi_1, \phi_2,\dots,\phi_d)$ denote the SHAP explanation vector for $\mathbf{x}$, and let $E_{\text{LIME}}(f,\mathbf{x}) = (\ell_1,\ell_2,\dots,\ell_d)$ denote the LIME explanation. We aggregate these two explanations into a single vector as  follows
\begin{equation}
\label{eq:centagg}
    E_{\text{agg}}(f,x) = \frac{1}{2}\big(E_{\text{SHAP}}(f,x) + E_{\text{LIME}}(f,x) \big)
\end{equation}
This choice is justified in the results of \cite[Prop. 2]{bhatt2020evaluating}, which shows that an average (centroid) can yield a more stable explanation than either component. 


\subsubsection{Perturbation}
To evaluate the stability of the explanation, we generate perturbed versions of the input and observe how the explanations change of $\mathbf{x}$ change. Formally, for a given data point $\mathbf{x}\in\mathbb{R}^d$, we define a random perturbation $\tilde{\mathbf{x}}$ drawn from a distribution $P_{\mathbf{x}}$ in the neighborhood of $\mathbf{x}$, which incorporates both Gaussian noise and feature masking:
\begin{itemize}
    \item \textbf{Gaussian noise}: For each feature, let $\boldsymbol{\epsilon}\sim \mathcal{N}(\mathbf{0},\sigma^2 I)$ (a $d$-dimensional Gaussian with variance $\sigma^2$ per feature). Then a perturbed instance is $\mathbf{x}+\boldsymbol{\epsilon}$. This simulates natural fluctuations in input that should not cause significant changes in explanation \cite{chen2022makes}.
    \item \textbf{Feature masking:} To simulate missing or uninformative features, we randomly replace some features with a baseline value (e.g. mean or zero). Let $\mathbf{x}^{(0)}$ be a reference baseline (e.g., the dataset mean). We introduce a random mask vector $M\in\{0,1\}^d$ where $M_j=0$ refers to mask feature $j$. For each feature $j$, with probability $p_j$ we set $M_j=0$ (mask it), and with probability $1-p_j$ we set $M_j=1$ (keep it). Then
 \[
    \tilde{x}_j = 
\begin{cases}
x^{(0)}_j, & M_j = 0~\text{ (feature $j$ masked)}\\
x_j + \epsilon_j, & M_j = 1~\text{ (feature $j$ noised).}~,
\end{cases}T
    \]
   \end{itemize}

In vector form, one draws from the perturbation distribution as $\tilde{\mathbf{x}} = M \odot (\mathbf{x} + \boldsymbol{\epsilon}) + (\mathbf{1}-M)\odot \mathbf{x}^{(0)}$, where $\odot$ denotes the element-wise product. 
Importantly, we restrict perturbations to a small neighborhood so that $f(\tilde{\mathbf{x}}) \approx f(\mathbf{x})$, i.e., the model prediction does not change significantly 

\subsubsection{Stability}
Using the above perturbation, we quantify the stability of the explanation by measuring how much the explanation changes. Let $E(f,\mathbf{x}) = E_{\text{agg}}(f,\mathbf{x})$ denote the aggregated explanation vector for the original input $\mathbf{x}$, and $E(f,\tilde{\mathbf{x}})$ the explanation for a perturbed input $\tilde{\mathbf{x}}$. 
We measure the distance between the original and perturbed explanations as
\[    d_{1}\!\big(E(f,\mathbf{x}),\,E(f,\tilde{\mathbf{x}})\big) \;=\; \sum_{j=1}^d \Big| E_j(f,\mathbf{x}) - E_j(f,\tilde{\mathbf{x}})\Big|.
\]
A small distance means the explanation is robust to the perturbation; however, a large distance means the feature attributions changed substantially, indicating instability. We define the explanation instability at $\mathbf{x}$ as the expected change in distance under random perturbations:
\begin{equation}
\label{eq:instab}
  \Delta(\mathbf{x}) \;=\; \mathbb{E}_{\tilde{\mathbf{x}}\sim P_{\mathbf{x}}}\Big[\,\|\,E(f,\mathbf{x}) - E(f,\tilde{\mathbf{x}})\|_{1}\,\Big],~
\end{equation}
where the expectation is over the perturbation distribution $P_{\mathbf{x}}$. 
In practice, we approximate $\Delta(\mathbf{x})$ by averaging $||E(f,\mathbf{x})-E(f,\tilde{\mathbf{x}})||_1$ over several sampled perturbations $\tilde{\mathbf{x}}$. A lower $\Delta(\mathbf{x})$ means a more stable explanation, 
while a higher value means that the explanation is sensitive to small input changes.

For convenience, we also define a stability score $S(\mathbf{x})$ such that a higher $S$ means higher stability. For example, $S(\mathbf{x}) = \exp(-\Delta(\mathbf{x}))$ or $S(\mathbf{x}) = 1/(1+\Delta(\mathbf{x}))$ would decrease monotonically as $\Delta$ increases. 


\subsubsection{Group-wise Aggregation of Stability Scores}
Next we extend the stability measure to evaluate group fairness. Suppose we have a sensitive attribute (e.g., gender or race) that partitions the dataset into $K$ disjoint groups $G_1, G_2, \dots, G_K$. We first compute the stability measure for each individual data point as above. Then, for each group $G_i$, we aggregate the stability scores of instances by taking the average over all group members. Formally, let $X_{G_i} = \{\mathbf{x} \in X : \text{group}(\mathbf{x}) = G_i\}$ be the set of inputs belonging to group $i$. We define the group stability (or average stability) for group $G_i$ as:
\[
S_i \;=\; \frac{1}{|X_{G_i}|} \sum_{\mathbf{x}\in X_{G_i}} S_i (\mathbf{x})~,
\]
where $S_i (\mathbf{x})$ is the stability score for $\mathbf{x}$. 


\subsubsection{GESD Metric}
Finally, GESD is computed to quantify how different the stability scores are across the sensitive groups. In the simplest case of two groups \big(e.g., $G_0$ (male) and $G_1$ (female)\big), the disparity can be measured as the absolute difference in their average stability scores:
\begin{equation*}
\text{GESD}_{2\text{-groups}} \;=\; \big|\,S_0 \,-\, S_1\,\big|~,
\end{equation*}
where $S_0$ and $S_1$ are the mean stability measures for group 0 and group 1, respectively. 
This aligns with common fairness metrics that use differences in performance or error rates between groups as measures of bias or disparity.

For the general case of $K>2$ groups, we use the mean squared deviation of the group stability scores ${S_1, S_2, \dots, S_K}$ from their overall mean $\bar{S} = \frac{1}{K}\sum_{i=1}^K S_i$. We define:
\begin{equation}
\label{eq:gesd}
\text{GESD}_{K\text{-groups}} \;=\; \frac{1}{K} \sum_{i=1}^K \Big(S_i \,-\, \bar{S}\Big)^2,
\end{equation}
which is the variance of the stability scores across the groups. 

\section{Fairness, Explainability, and Utility}
\label{sec:feu}
In line with the novel explainability metric proposed above, and due to the exceptional competence of evolutionary algorithms in solving combinatorial optimization problems \cite{rehman2022fair, wang2024generating}, this work solves a multiobjective optimization problem focused on balancing Fairness, Explainability, and Utility (FEU).
We frame our FEU as an unconstrained three-objective optimization problem, solved using the Non-dominated Sorting Genetic Algorithm II (NSGA-II) \cite{deb2002fast}. In our setup, each individual in the population encodes a particular model configuration, including both model hyperparameters. For example, a solution $\mathbf{x}$ can be represented by $\mathbf{x} = (\gamma, \rho, \tau)$, consisting of the learning rate $\gamma$, the epoch $\rho$ for model training, and a decision threshold $\tau$ for positive classification.


Let $\mathbf{x}$ denote the vector of decision variables.
We define three objective functions $
 \mathcal{L}_f(\mathbf{x}), 
\mathcal{L}_e(\mathbf{x}),\mathcal{L}_u(\mathbf{x})$ corresponding to fairness, and explainability, and utility respectively. Each objective is formulated as a quantity to be minimized (note that $\mathcal{L}_u$ (utility) is maximized but was converted to minimization through negation). The multi-objective optimization problem is then stated as: 
\begin{equation}
\label{eq:moo}
\min_{\mathbf{x} \in \mathcal{X}} ;; \big((\mathcal{L}_f(\mathbf{x}), 
\mathcal{L}_e(\mathbf{x}),- \mathcal{L}_u(\mathbf{x})) \big)~, \end{equation}
where $\mathcal{X}$ is the space of all valid model configurations. 
We next describe the objectives in detail:

\begin{enumerate}
    \item \textbf{Fairness} For fairness, we adopt the Demographic Parity (DP) criterion; however, any outcome-based criterion can be used. Let $G$ be a specific binary sensitive attribute (e.g., gender or race), with values 0 and 1 for the two demographic groups. Demographic Parity requires the positive prediction rate of the model to be independent of $G$. We quantify the deviation from this fairness ideal by the absolute difference in positive prediction rates between the groups \cite{chouldechova2017fair}:
\[
\mathcal{L}_f(\mathbf{x}) = \big| P (\hat{Y}=1|G=0) - P(\hat{Y}=1|G=1) \big|
\]
where $\hat{Y}$ is the model predicted label. This DP difference $\mathcal{L}_f(\mathbf{x})$ is a non-negative number, with 0 indicating perfect fairness.  We aim to minimize $\mathcal{L}_f(\mathbf{x})$, yielding a fair model.

\item \textbf{Group-level Explanation Disparity (GESD)} The second objective ($\mathcal{L}_e$) is our metric, defined in Section \ref{sec:gesd}. The goal is for GESD to quantify the difference in explanation disparity between groups within protected attributes. 

\item \textbf{Utility} For utility, we use the Area Under the Receiver Operating Curve (AUC). Formally, if $N_{+}$ and $N_{-}$ are the number of positive and negative examples in the data, and $f(x)$ is the model score for instance $x$, then the AUC of $f$ corresponds to the probability that a random positive example receives a higher score than a random negative example. 
\[
\mathcal{L}_u(\mathbf{x})= \frac{1}{N_+N_-} \sum_{i:y_i=1} \sum_{i:y_j=0} \mathcal{I}(f(\mathbf{x}_i) > f(\mathbf{x}_j))
\]
\end{enumerate}

Together, we seek to minimize objectives 1 and 2 (fairness and explainability) and maximize objective 3 (AUC). We normalize these objectives to comparable scales to aid the optimization. Formally, the algorithm seeks to approximate the Pareto-optimal set ${\mathbf{x}^*}$ for the vector objective in Equation \eqref{eq:moo}.

We solve the above multi-objective problem using NSGA-II. NSGA-II iteratively evolves a population of candidate solutions toward better trade-offs among the objectives, using principles of natural selection (based on Pareto dominance) and genetic variation (crossover and mutation). We choose NSGA-II because it can efficiently approximate the Pareto front (the set of non-dominated solutions) in a single run.
In NSGA-II, each population member corresponds to a model configuration $\mathbf{x}$ as defined earlier. We start by initializing an initial population $P_0$ of $N$ random solutions, sampling each hyperparameter and weight from a predetermined range. We evaluate every solution $\mathbf{x}$ in the population on the three objectives 
described above (which entails training a model with the given hyperparameters and computing its DP difference, GESD, and AUC on a test dataset).


\section{Experimental Evaluation}
\label{sec:exp}
In this section, we report on several experiments to evaluate our proposed procedural metric (GESD) and test the effectiveness of our proposed FEU optimization process. Specifically, we answer the following research questions: 
\begin{itemize}
    \item RQ1: How well can FEU mitigate bias related to traditional outcome-oriented and procedural-oriented fairness measurement?
    \item RQ2: How well can FEU balance different categories of objectives compared to other baselines?
    \item RQ3: How well does GESD measure procedural-oriented bias in linear and nonlinear models?
\end{itemize}

\subsection{Experimental Settings}
\subsubsection{Dataset:} We evaluated the proposed framework on four real-world datasets: German Credit \cite{asuncion2007uci}, Recidivism \cite{jordan2015effect}, Math, and Portuguese \cite{cortez2008using}, which are state-of-the-art datasets for measuring algorithmic bias.
\subsubsection{Baselines:} 
Given that no prior work has explicitly considered explanation fairness, we compare our work with three fair learning methods: Adversarial learning \cite{zhang2018mitigating} (ADL), Reduction (Red) \cite{agarwal2018reductions}, and Reweighing (RW) \cite{kamiran2012data}. Note that ADL uses a neural network (NN) as the base model, while reduction and reweighing use logistic regression as their base model.

\subsubsection{Evaluation Criteria:} We evaluate model performance from three perspectives: (1) model utility: AUC and F1; (2) outcome-oriented fairness: $\Delta_{DP}$ and $\Delta_{EOD}$  \footnote{$\textsc{EOD} = P\{\hat{Y}=1|G=0, Y =y \}= P\{\hat{Y}=1|G=1, Y=y\}, y\in\{0,1\}  $} \cite{hardt2016equality}; and (3) procedural-oriented fairness: the GESD metric given in Equation (\ref{eq:gesd}). 

\subsubsection{Setup:} For GESD, we used 100 samples with SHAP and LIME because some of our datasets had fewer than 1000 examples. We used a feedforward neural network with two hidden layers as the underlying model for FEU. With NSGA-II, wee used Simulated Binary Crossover (SBX) and Polynomial Mutation (PM), where the crossover probability was set to $0.9$, which means that $90\%$ of the selected pairs undergo recombination. The SBX distribution index was fixed at $\eta = 15$. For mutation, we set $\eta = 20$, which controls the spread of the offspring during crossover or mutation. FEU has three hyperparameters, given earlier, i.e., $(\gamma,\rho,\tau)$.

\subsubsection{Model selection:} To select one solution from the Pareto fronts produced by our multi-objective optimization, we considered four popular model-selection methods: Elbow, Linear Scalarization, Chebyshev, and Hypervolume \cite{wei2022fairness}. The Elbow method selects the point at which improvement along one objective begins to yield diminishing returns along the others. Linear Scalarization converts the three-objective optimization problem into a single-objective one by computing a weighted sum $( S(x) = \sum_{i=1}^{3} w_i f_i(x) )$, where $ ( f_1, f_2, f_3 )$ denote normalized AUC, DP, and GESD, respectively. While simple, scalarization can underrepresent non-convex regions of the frontier and can miss optimal solutions in non-linear trade-offs.

We also employ the Chebyshev scalarization, which measures the maximum weighted deviation from the ideal point and therefore captures worst-case performance across objectives. Formally, letting $\mathbf{z}^\star$ denote the ideal vector and $\mathbf{f}(x) = (f_1(x), f_2(x), f_3(x))$ the objective values, the Chebyshev score is
\[
T(x) = \max_i\ w_i \lvert f_i(x) - z_i^\star \rvert,
\]
and the solution selected is $x^\star = \text{argmin}_x\ T(x)$. This criterion favors solutions that avoid extreme degradation in any single objective, thereby providing balanced and robust selection properties. The fourth method we considered was Hypervolume, which measures the volume dominated by each point in the Pareto set relative to a reference point; a larger hypervolume corresponds to stronger performance across the three criteria. 

Across all experiments, we assign weights of 0.4 to AUC and 0.3 each to fairness and explainability. These values are consistent with earlier multi-objective studies that discourage any single metric from dominating the trade-off space, while providing a slight preference for predictive performance.
Based on both empirical stability and theoretical grounding, we adopt the Chebyshev selection as our final method. 

\begin{table}[t!]
\centering
\caption{Comparison of FEU and baselines across datasets. 
\label{tbl:feu}
$\uparrow$ higher is better, $\downarrow$ lower is better. 
Bold indicates best performance.}
\label{tab:results}
\resizebox{\columnwidth}{!}{
\begin{tabular}{|l|c|c|c|c|c|c|}
\hline
\textbf{Data} & \textbf{Metric} & \textbf{FEU} & \textbf{ADL} & \textbf{Red.} & \textbf{RW} \\
\hline
\multirow{5}{*}{German}
 & AUC $\uparrow$ & \textbf{0.72} & 0.71 & 0.63 & 0.69 \\
 & F1 $\uparrow$ & \textbf{0.77} & 0.72 & 0.71 & 0.75 \\
 & DP $\downarrow$ & \textbf{0.12} & 0.14 & 0.15 & 0.24 \\
 & EOD $\downarrow$ & 0.14 & \textbf{0.13} & 0.17 & 0.27 \\
 & GESD $\downarrow$ & \textbf{0.01} & 0.05 & 0.03 & \textbf{0.01} \\
\hline
\multirow{5}{*}{Por}
 & AUC $\uparrow$ & \textbf{0.94} & 0.93 & 0.88 & 0.92 \\
 & F1 $\uparrow$ & 0.69 & \textbf{0.85} & 0.75 & 0.85 \\
 & DP $\downarrow$ & 0.15 & 0.17 & \textbf{0.02} & 0.08 \\
 & EOD $\downarrow$ & \textbf{0.25} & 0.43 & 0.27 & 0.26 \\
 & GESD $\downarrow$ & 0.03 & 0.05 & 0.02 & \textbf{0.01} \\
\hline
\multirow{5}{*}{Math}
 & AUC $\uparrow$ & 0.91 & \textbf{0.94} & 0.86 & 0.93 \\
 & F1 $\uparrow$ & 0.77 & \textbf{0.83} & 0.82 & 0.82 \\
 & DP $\downarrow$ & 0.16 & \textbf{0.03} & 0.04 & 0.04 \\
 & EOD $\downarrow$ & \textbf{0.07} & 0.17 & 0.23 & 0.06 \\
 & GESD $\downarrow$ & 0.004 & 0.07 & 0.003 & \textbf{0.002} \\
\hline
\multirow{5}{*}{Recid.}
 & AUC $\uparrow$ & \textbf{0.99} & 0.94 & 0.80 & 0.93 \\
 & F1 $\uparrow$ & \textbf{0.98} & 0.88 & 0.89 & 0.76 \\
 & DP $\downarrow$ & 0.06 & 0.12 & \textbf{0.03} & 0.05 \\
 & EOD $\downarrow$ & \textbf{0.001} & 0.03 & 0.02 & 0.006 \\
 & GESD $\downarrow$ & 0.01 & 0.02 & 0.05 & \textbf{0.007} \\
\hline
\end{tabular}}
\end{table}

\subsection{Performance Comparison}
To answer RQ1, we evaluate FEU and other fairness models on four real-world datasets and report the results for utility, outcome-oriented fairness, and procedural-oriented fairness. The results in Table \ref{tbl:feu} show varying performance across the algorithms. From Table \ref{tbl:feu}, we can infer the following:
\begin{itemize}
    \item FEU slightly outperforms other methods in three of the four datasets on utility, and on the Math dataset, FEU's utility is still competitive. This shows that 
    much of the utility was not sacrificed for other objectives. Also, we see that models with non-linear decision boundaries, like ADL, were able to generate higher utility than Red. and RW, which used logistic regression.
    \item FEU also outperformance the other methods on the outcome-oriented fairness metrics, DP and EOD. 
    However, in some cases FEU lost out on DP in favor of EOD. This occurred because of inherent trade-offs between multiple fairness metrics \cite{tarzanagh2023fairness}.
    \item On the procedural-oriented metric (GESD), FEU shows lower performance in comparison to Red. and RW. The results are not surprising, as FEU makes many complex decisions, which can lead to explanation disparities between subgroups within a protected category. The results further demonstrate the need to explain the decisions received by each subgroup. GESD is superior to \cite{zhao2023fairness} because it does not rely on model prediction. This demonstrates that it is possible for a model to have higher utility and higher fairness but greater explanation disparity, which is why it is necessary to optimize for these three criteria before deciding on the final model.    
\end{itemize}

\subsection{Trade-off Between Objectives}
\begin{figure}[t!]
    \centering
    \includegraphics[width=1.0\linewidth]{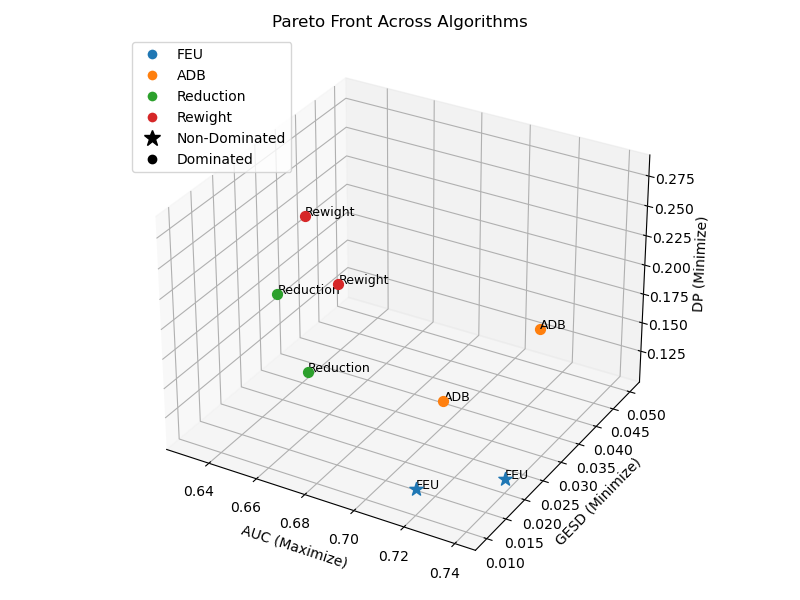}
    \caption{Pareto optimal solutions on German Credit dataset}
    \label{fig1}
\end{figure}

To demonstrate the effectiveness of each algorithm in balancing fairness, explainability, and utility, we compare the Pareto sets on the German Credit dataset in Figure \ref{fig1}. The plotted results demonstrate that Red. and RW, both of which employ logistic regression as their base classifiers, consistently yield dominated solutions. These methods often sacrifice significant predictive utility (measured by AUC) to achieve marginal gains in fairness, a trade-off that limits their applicability in real-world high-stakes scenarios.
However, FEU distinctly generates Pareto-optimal (non-dominated) solutions, demonstrating its ability to balance the three crucial objectives effectively. 

Interestingly, while ADL achieves competitive utility and outcome-oriented fairness, it underperforms in procedural-oriented fairness, exhibiting the highest GESD values. FEU also shows this trend of being outperformed in GESD in other datasets by Red. and RW, which emphasizes an essential limitation of outcome-based metrics---they alone are insufficient for identifying and mitigating disparities related to model explainability and procedural fairness. Consequently, GESD is invaluable as a complementary metric to provide insights into algorithmic decision-making processes and highlighting disparities otherwise hidden by traditional fairness metrics. 

\subsection{Comparing GESD Across Models}
\begin{figure}[t!]
    \centering
    \includegraphics[width=1.05\linewidth]{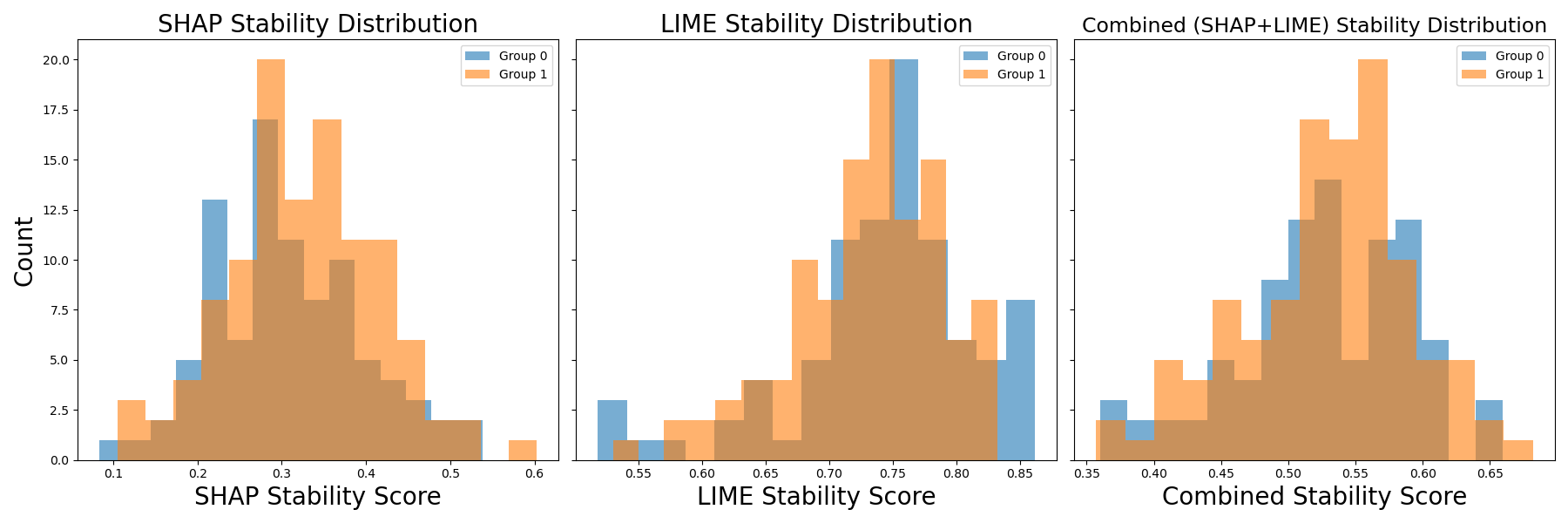}
    \caption{Reweighing Stability on Recidivism Dataset}
    \label{fig5}
\end{figure}

\begin{figure}[t!]
    \centering
    \includegraphics[width=1.05\linewidth]{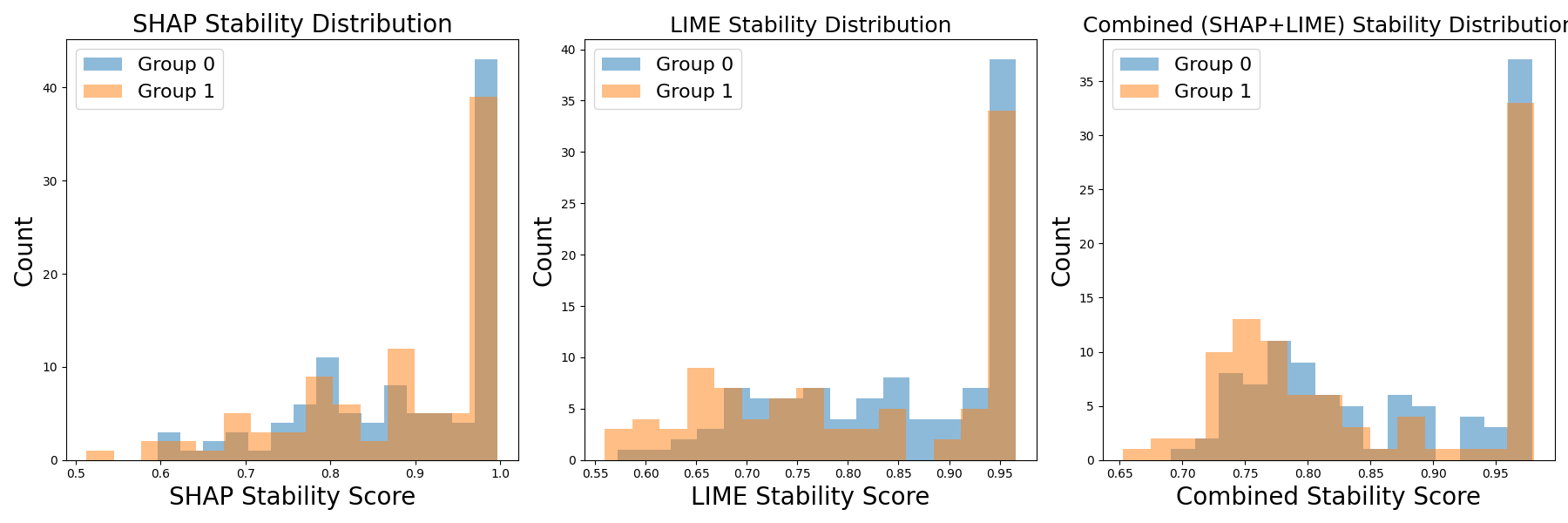}
    \caption{FEU Stability on Recidivism Dataset}
    \label{fig6}
\end{figure}

To test how well GESD measures procedural bias across different models, particularly linear and nonlinear models, we conducted an experiment by selecting two methods: FEU and RW. We selected these models because FEU used a neural net, which is nonlinear, while RW used logistic regression, which is linear. 
We plot both the RW and FEU SHAP, LIME, and Combined distributions on the Recidivism dataset in Figures \ref{fig5} and \ref{fig6}, respectively. Also, to determine if the explanation disparities were significantly different, we performed Mann-Whitney U tests of each pair of $G_0$ and $G_1$ on SHAP, LIME, and Combined, and we report the p-values in Table \ref{tab2}. 

The results in Table \ref{tab2} show important differences in how procedural bias manifests across explainers and model classes. For RW, SHAP and the combined explainer do not indicate statistically significant disparities in explanation between groups, whereas LIME does. This demonstrates that explanation disparities in linear models can be small but still detectable, depending on the explainer used. In contrast, the FEU neural network exhibits stronger and more consistent disparities across explainers, with LIME and the combined explainer showing significant group differences. Interestingly, SHAP alone does not reveal disparity for FEU, highlighting that individual explainers vary in sensitivity. Together, these results support prior observations \cite{dai2022fairness} that complex, non-linear models tend to generate less stable or more sensitive explanations across subgroups.

These results also validate the intuition behind GESD, which is that a model can satisfy outcome-oriented fairness while still exhibiting explanation-oriented disparities. The variation in significance across SHAP, LIME, and their combination further demonstrates the importance of using multiple explainers when assessing procedural fairness. Because some explainers may fail to detect instability that others reveal, aggregating explainers helps to capture disparities robustly and consistently across model classes and datasets.

\begin{table}[t!]
    \centering
    \begin{tabular}{|c|c|}
        \hline
        \textbf{Comparison} & \textbf{P-value} \\ \hline
        Reweighing\_SHAP & 0.96\\ \hline
        Reweighing\_LIME & 0.01 \\ \hline
        Reweighing\_Combine & 0.27 \\ \hline
        FEU\_SHAP &  0.25 \\ \hline
        FEU\_LIME  & 0.008\\ \hline
        FEU\_Combine & \textbf{0.002} \\ \hline
    \end{tabular}
    \caption{Reweighting and FEU GESD  on Recidivism dataset}
    \label{tab2}
\end{table}

\section{Discussion}
\label{sec:disc}

The results of our experiments demonstrate that FEU outperforms other algorithms on three of the four data sets in terms of utility. This performance shows that FEU is not trading accuracy for performance, as seen in ADL, Red., or RW, which is consistent with results in other work \cite{rehman2022fair, wang2024generating}. 
The outcome-oriented fairness results (DP and EOD) also show that FEU exhibits competitive performance, though it appears the results favor EOD over DP. This is due to an inherent trade-off between different fairness metrics and the difficulty in optimizing multiple fairness metrics. Overall, however, the outcome-oriented results show less bias on both metrics on all the datasets with our method. For the procedural-oriented metrics, FEU yields a higher (but statistically insignificant) GESD in comparison to Red. and RW on three out of four datasets. Note, however, that the three datasets are small in terms of the number of instances, and neural network models often require more samples to make good predictions. Also, since the neural network makes complex decisions in comparison to logistic regression, this can lead to explanation disparity in FEU. This can be verified as ADL, which also uses a neural net, also consistently yielded higher GESD in comparison to other models. However, the results of GESD for FEU and ADL both show a lower overall explanation quality.

From the solutions shown in Figure \ref{fig1}, we demonstrated that the results of FEU are non-dominated across the three objectives on the German Credit dataset; however, the results are slightly different on other datasets. This is due to the fact that the other datasets are extremely biased, and our FEU algorithm did not contain any fairness intervention method as in ADL, which uses adverserial learning to debias the data, or in RW, which creates a weight for each sample based or the amount of bias in the dataset. This is an area of future work, where we intend to incorporate bias-mitigation techniques in FEU to make it more robust to extremely biased datasets. 

GESD yielded different performance across the various models. These results align with the overall goal of this paper, which is to develop a novel, robust, procedural-oriented fairness metric for opaque models. This metric determines whether the explanation generated by the model is the same across each subgroup of a protected category. 
An interesting observation from Figures \ref{fig5} and \ref{fig6} is highlighting the varying explanations generated by SHAP and LIME. This pattern was observed in all datasets and across all algorithms. The results validate the purpose of aggregating two explainers to produce GESD. These inconsistencies have been noted in several results in the literature \cite{slack2021reliable}. The aggregation used in GESD helps overcome this shortcoming of the explainer, as we see in the results in Table \ref{tab2}. If we use only SHAP for GESD, we would not see any difference between the explanations generated for each subgroup, but the aggregation exposes the disparity in the explanations generated for each subgroup.


\section{Conclusion and Future Work}
\label{sec:conc}

We introduced GESD, a procedural fairness metric designed to measure the quality of explanation disparity between subgroups of a protected category. The potential of GESD lies in its ability to measure disparity between the subgroups of protected categories without relying on model outcomes or predictions.  We also made GESD more robust to inconsistencies from the explainer by aggregating two explainers. This regularizes performance between the explainers to reduce disparity regardless of the performance of the explainer. We later combined GESD with AUC and DP, and used FEU to perform robust multi-objective optimization of these three metrics to select the best model that balances them. The effectiveness of FEU was demonstrated through its consistent ability to achieve the best, non-dominated results in fairness and utility. Also, the effectiveness of GESD was demonstrated through its ability to expose procedural bias, whether in linear or non-linear models. 

In the future, we plan to extend GESD to multiple protected categories (e.g., race and gender) to address intersectional bias.  Another plan for GESD is to expand it to consider the aggregated objectives (currently, for LIME and SHAP) from a truly multi-objective perspective. 
This will help in mitigating situations where GESD is high for a particular subgroup. Turning this GESD into a multi-objective problem will convert the optimization of fairness, explainability, and utility into a many-objective problem (4 or more objectives). 
In order to tackle the many objective problems, we plan to shift to an alternative MOEA method, such as NSGA-III \cite{deb2013evolutionary,Deb6595567}, which employs a reference point-based approach to address the challenges of handling many objectives, . 


\balance
\bibliographystyle{IEEEtran}
\bibliography{aaai25}

\end{document}